\documentclass{article}

\usepackage{microtype}
\usepackage{graphicx}
\usepackage{subfigure}
\usepackage{booktabs}
\usepackage{hyperref}

\usepackage[accepted]{icml2024}

\usepackage{amsmath}
\usepackage{amssymb}
\usepackage{mathtools}
\usepackage{amsthm}

\usepackage[capitalize,noabbrev]{cleveref}

\usepackage[textsize=tiny]{todonotes}

\icmltitlerunning{Projectable Models: One-Shot Generation of Small Specialized Transformers from Large Ones}

\theoremstyle{plain}

\theoremstyle{definition}

\theoremstyle{remark}

\def\R{\mathbb{R}}
\def\Z{\mathbb{Z}}

\def\E{\mathbb{E}}

\renewcommand{\vec}[1]{\boldsymbol{#1}}

\newcommand{\mc}[1]{\mathcal{#1}}

\renewcommand{\vec}[1]{\boldsymbol{#1}}

\renewcommand{\comment}[1]{{\color{BrickRed} #1}}
\newcommand{\hidecomment}[1]{}

\def\task{\vec{t}}
\def\tasks{T}
\def\digit{d}
\def\image{\vec{i}}

\newcommand{\op}[1]{\hat{\mathrm{#1}}}
\newcommand{\mat}[1]{\mathrm{#1}}
\newcommand{\model}[1]{\mc{M}_{#1}}
\newcommand{\oper}[1]{\op{O}_{#1}}
\newcommand{\cmodel}[1]{\tilde{\mc{M}}_{#1}}
\newcommand{\proj}[1]{\mc{P}_{#1}}
\def\jointproj{\tilde{\mc{P}}}

\def\msizes{\{1/2,1/4,1/8\}}

\def\pmap{\rho}
\def\ffamily{\mc{F}}
\def\synth{\mbox{\textsc{SynthMNIST}}}
\def\mnist{\textsc{MNIST}}
\def\gpt{\textsc{GPT-2}}
\def\imagegpt{\textsc{ImageGPT}}
\def\imagenet{\textsc{ImageNet}}
\def\simclr{\textsc{SimCLRv2}}

\begin{document}

\twocolumn[
\icmltitle{Projectable Models: One-Shot Generation of Small Specialized Transformers from Large Ones}

\icmlsetsymbol{equal}{*}

\begin{icmlauthorlist}
\icmlauthor{A. Zhmoginov}{goog}
\icmlauthor{J. Lee}{goog}
\icmlauthor{M. Sandler}{goog}
\end{icmlauthorlist}

\icmlaffiliation{goog}{Google DeepMind}

\icmlcorrespondingauthor{Andrey Zhmoginov}{azhmogin@google.com}

\icmlkeywords{Transformers, image generation, model compression, weight generation, hypernetworks}
\vskip 0.3in
]

\printAffiliationsAndNotice{\icmlEqualContribution}

\renewcommand{\comment}[1]{}

\begin{abstract}
    Modern Foundation Models (FMs) are typically trained on corpora spanning a wide range of different data modalities, topics, downstream tasks.
    Utilizing these models can be very computationally expensive and is out of reach for most consumer devices.
    Furthermore, most of the broad FM knowledge may actually be irrelevant for a specific task at hand.
    Here we explore a technique for mapping parameters of a large Transformer to parameters of a smaller specialized model.
    By making this transformation task-specific,
    we aim to capture a narrower scope of the knowledge needed for performing a specific task by a smaller model.
    We study our method on image modeling tasks, showing that performance of generated models exceeds that of universal conditional models.
\end{abstract}

\vspace{-2em}
\section{Introduction}

    Transformer-based generative models have recently shown a remarkable success in modeling complex data distributions across a wide spectrum of modalities including images, audio and language.
In the language domain, Large Language Models (LLMs) became an extremely powerful tool demonstrating impressive performance across a large scope of language tasks.
These models are typically very computationally complex to train and run and require vast amounts of text data that encompass a wide range of topics, facts and downstream applications.
The generality of the resulting LLMs can be extremely advantageous, but also prove redundant or even detrimental in narrow applications and in specialized tasks \cite{Raffel2020exploring}.

Here we propose an approach to generating a task-specific Transformer model of a smaller size from a larger Transformer.
In our primary setup, similar to that explored in other related publications (see Appendix~\ref{sec:related}), we maintain a functional universal large Foundation Model (FM) and use it to generate smaller specialized task-dependent Transformers with an emphasis on multi-task learning and personalization scenarios.
Following the HyperNetwork approach \cite{HaDL16}, we generate the specialized models on the fly from the task specification without the need for any fine-tuning.

We study this idea experimentally in the visual domain adopting \imagegpt{} approach \cite{imagegpt}, in which images are represented as sequences of discrete tokens and an autoregressive Transformer model is used to model this sequence distribution.
Here we compare performance of task-specialized generated small Transformer models with universal counterparts of the same size.
We also explore zero-shot generalization and knowledge transfer across different tasks.

\section{Method}
\label{sec:method}

    \def\token{\vec{x}}
\def\act{\vec{z}}
\def\mapdown{\op{P}}
\def\mapup{\op{Q}}
\def\matdown{\mat{P}}
\def\matup{\mat{Q}}

In the following, we outline our approach, where we use large Transformer models for generating smaller task-specific Transformers.
We simultaneously view this as (a) a model specialization technique, where we can produce a small Transformer best suited for a particular task; (b) model compression technique where a single large model can be used for producing a variety of models of different sizes, and finally (c) a modular network approach, where individual generated task-specialized model implicitly share knowledge via a single large model.
A more detailed discussion of weight space manifolds and associated model specialization techniques can be found in Appendix~\ref{app:perspective}.

\begin{figure}
  \centering
  \includegraphics[width=0.45\textwidth]{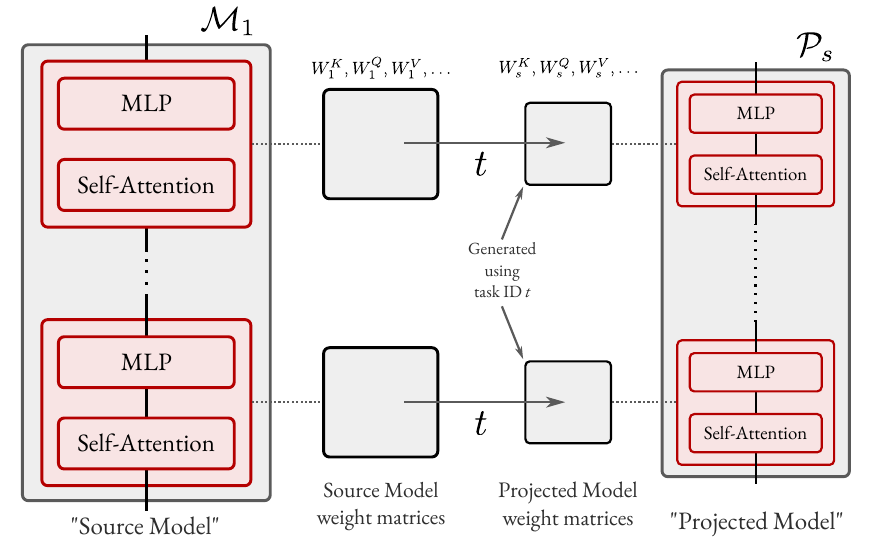}
  \caption{
    Model diagram.
    Given the task ID $\task$, the weights of the large ``source model'' $\model{1}$ (Transformer with the embedding size $d$) are mapped to the weights of a smaller ``projected model'' $\proj{s}$ (Transformer with the embedding size $ds$).
    The loss of $\proj{s}$ on the input sequence is used to train the ``projection operator'' and tune $\model{1}$, while a good performance of $\model{1}$ can be maintained by continuing to train it with the original objective.
  }
  \label{fig:model}
\end{figure}

\subsection{Matrix Generators}
    \label{sec:generators}

    In the Transformer architecture, the embedding size typically defines the sizes of all self-attention and MLP weight matrices.
    Choosing some maximum embedding size $d$, we define a family of Transformer models $\{\model{s}\}_{s\in S}$ with the embedding size $ds$.
    We typically choose $s=2^{-k}$ with integer $k\ge 0$ assuming that $d/2^{k}$ is integer for all models of interest.
    We generally assume that each $\model{s}$ is a separate model pretrained on some input data distribution $p(\token)$.

    Given pretrained $\model{1}$ how can we generate a smaller Transformer with architecture of $\model{s}$ and the embedding size $ds$?
    One natural way of doing this is to map each linear operator\footnote{all dense layers in self-attention and MLP operators} $\oper{1}^\ell(\act):=\mat{W}^\ell_1 \act + \vec{b}^\ell_1$ from $\model{1}$ to a corresponding linear operator\footnote{In the following we use a hat to denote operators ($\op{X}$) and roman font for matrices ($\mat{X}$).} $\oper{s}^\ell(\act):=\mat{W}^\ell_s \act + \vec{b}^\ell_s$ in the small model.
    Assuming that this transformation is linear, its most general form is simply
    \begin{gather*}
        (\mat{W}_s)_{ij} := \sum_{k,l} \mat{V}^W_{ijkl} (\mat{W}_1)_{kl} + \mat{V}^B_{ij}, \\
        (\vec{b}_s)_i := \sum_{k} \mat{U}^W_{ik} (\vec{b}_1)_{k} + \mat{U}^B_{i}.
    \end{gather*}
    
    \paragraph{Projections.}
    One natural choice of $(\mat{V}^W,\mat{V}^B,\mat{U}^W,\mat{U}^B)$ comes from analyzing linear transformations of the embedding space.
    Specifically, given two linear maps $\mapup:\R^{ds} \to \R^{d}$ and $\mapdown:\R^{d} \to \R^{ds}$ mapping embeddings between embedding spaces of two models, we can naturally define $\oper{s}$ via $\oper{s}(\act) \equiv \mapdown (\oper{1}(\mapup\act))$.
    In other words, the action $\oper{s}(\act)$ is defined as a sequence of several steps: (a) mapping $\act$ to a higher-dimensional embedding space via $\mapup \act$, (b) acting on this activation with $\oper{1}$ and finally (c) projecting the result down to the original low-dimensional embedding space with $\mapdown$.
    Ignoring biases, we can define $\mapup(\act)=\matup \act$ and $\mapdown(\act)=\matdown \act$ with $\matup$ and $\matdown$ being two matrices, thus simply obtaining:
    \begin{gather}
        \label{eq:Wb}
        \mat{W}_s = \matdown \mat{W}_1 \matup, \qquad \vec{b}_s = \matdown \vec{b}_1.
    \end{gather}
    In the following, we mainly adopt this model.
    
\subsection{Task-Dependent Matrix Generators}
    \label{sec:generators}

    Our primary goal is generating small {\em task-specific} Transformer models, which requires that the generated weight matrices $\mat{W}_s$ depend on the task.
    In the following, we assume that each task is uniquely defined by a task identifier $\task\in \tasks$ with $\tasks$ being a finite-dimensional vector space of all possible tasks.
    Given $\task$, we then define a {\em projection operator} $\pmap_s$ that first maps $\task$ to $\mat{V}(\task)$ and $\mat{U}(\task)$ at every layer as discussed below and then uses these tensors and the large Transformer {\em source model} $\model{1}$ to generate the weights of a smaller {\em projected model} $\proj{s}=\pmap_s(\task;\model{1})$, in effect defining a function family $\ffamily(\rho_s,\model{1}) = \{ \pmap_s(\task;\model{1}) | \task \in \tasks \}$.
    
    If the tasks are specified explicitly as a part of the training dataset, we expect that $\pmap_s(\task;\model{1})$ performs a task $\task$ better than a universal task-agnostic model of the same architecture.
    On the other hand, if task identifiers are not provided, we could design synthetic tasks with the goal of making the resulting $\ffamily(\rho_s,\model{1})$ sufficiently rich to be useful for possible downstream tasks.

    \begin{figure}
          \centering
          \includegraphics[width=0.35\textwidth]{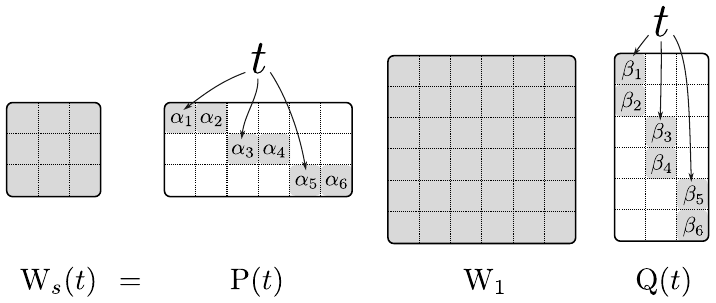}
          \caption{
            Projection of the source model weight $\mat{W}_1$ into the projected model weight $\mat{W}_s$.
            Here $\task$ is the task identifier and $\mat{P}(\task)$, $\mat{Q}(\task)$ are ``projection matrices'' whose diagonal elements are generated using shallow MLPs from $\task$.
          }
          \label{fig:mat-generation}
          \vspace{-1em}
    \end{figure}
    
    While there may be numerous choices of defining linear operators $\op{V}(\task)$ and $\op{U}(\task)$, here we adopt the {\em projection matrix} approach \eqref{eq:Wb} by learning $\mat{P}(\task)$ and $\mat{Q}(\task)$ matrices (see below).
    Our experiments suggested that the best way to generate high-quality specialized $\proj{s}$ required that the weights of $\model{1}$ are also tuned thus increasing the number of trained parameters and modifying a pretrained source model $\model{1}$ to be compatible with generated projections.
    We tuned $\model{1}$ and learned projection operators by optimizing both the original loss on $\model{1}$ to maintain its quality and the loss on the projected model $\proj{s}$ (see Eq.~\eqref{eq:loss} below).

    Since the projection matrix generator with many parameters can be susceptible to overfitting on a given task distribution, we propose a simple and low-parameter method of generating $\mat{P}\in \R^{(sdn)\times (dn)}$ and $\mat{Q}\in \R^{(dn)\times (sdn)}$.
    Namely, we chose to train a set of MLPs that given $\task$ generated two vectors $\vec{p}(\task)$ and $\vec{q}(\task)$ and chose:
    $\mat{P}_{ij} = p_{j}(\task) \delta_{i,\lfloor sj \rfloor}$ and $\mat{Q}_{ij} = q_i(\task) \delta_{j,\lfloor si \rfloor}$ (see Fig.~\ref{fig:mat-generation}).
    The resulting transformation $\mat{W}_s = \mat{P} \mat{W}_1 \mat{Q}$ can be seen as generating the output rows and columns by linearly combining $2^k$ rows and columns of the source matrix.
    While there may be many other approaches to generating weight matrices $\mat{W}_s$, we leave this question to be addressed in the future work.

\subsection{Optimization Objective}

    During training, we typically optimized the sum of the losses of the projected model $\proj{s}$ and the source model $\model{1}$:
    \def\vxi{\boldsymbol{\xi}}
    \begin{gather}
        \label{eq:loss}
        \E_{(\vxi,\task) \sim \mc{D}_*} \mc{L}_{\proj{s}(\task)}(\vxi) + w_{\rm src} \E_{\vxi \sim \mc{D}} \mc{L}_{\model{1}(\task)}(\vxi),
    \end{gather}
    where $\vxi=(\vec{x},\vec{y})$, $\mc{D}$ is the dataset used to train the Foundation Model $\model{1}$, $\mc{D}_*$ is the multi-task dataset for training projected models and $w_{\rm src}$ is an optional multiplier used to re-weight the importance of the Foundation Model objective.
    In autoregressive models, $\vec{y}$ is typically a shifted version of the sequence $\vec{x}$.

\section{Experiments}
\label{sec:experiments}

    In the following, we describe our experiments with model projections.
    All of our experiments were conducted in the image domain and followed the \imagegpt{} approach \cite{imagegpt} using two separate datasets: synthetic dataset \synth{} and a dataset based on \imagenet{}.
    
    \subsection{Experimental Setup: Datasets}
\label{sec:synth-mnist}

    \paragraph{\synth.}
    Our synthetic image dataset \synth{} was based on \mnist{} with each $32\times 32$ image generated by a deterministic function of a task identifier $\task\in \R^{18}$ specifying the background and $\digit$ encoding the index of the \mnist{} image to be overlaid on top of this texture.
    The vector $\task$ uniquely defined which of the three distinct types of the background textures to use and specified texture parameters such as scale, rotation, color and distortion.
    Treating $\digit$ as a hidden variable, the image distribution defining our multi-task dataset was given by $p(\image|\task)p(\task)=\sum_{\digit} p(\image|\task,\digit) p(\task) p(\digit)$.
    Some examples are illustrated in Figure~\ref{fig:sample}.
    For a more detailed discussion of the dataset see Appendix~\ref{app:models_datasets}.

    \begin{figure}[b]
      \centering
      \includegraphics[width=0.42\textwidth]{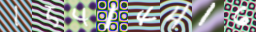}
      \caption{
        Examples of $32\times 32$ synthetic images generated from the task ID $\task$ encoding the background texture ($1$ of $3$ types of the texture, scale, rotation, colors, distortion, etc.) and the overlaid \mnist{} digit.
      }
      \label{fig:sample}
    \end{figure}

    \paragraph{\imagenet.}
    For our experiments with realistic images, we used $64\times 64$ RGB images from the \imagenet{} dataset \cite{imagenet}.
    The task identifier $\task$ associated with each image was chosen to be an embedding produced by a pretrained \simclr{} model \cite{simclrv2}.
    Specialized \imagegpt{} models are therefore trained to model a distribution of images with a given embedding $\task$.

\subsection{Experimental Setup: Model Architecture}
\label{sec:models}

    In our \synth{} experiments, all $32\times 32$ RGB input images were quantized by separately mapping each pixel into one of $512$ discrete tokens corresponding to one of the color clusters.
    These tokens were then flattened into $1024$-long sequences and finally modelled autoregressively using \gpt-style models \cite{Radford2019language}.
    Following \citet{imagegpt}, our experiments with $64\times 64$ RGB \imagenet{} images \cite{imagenet} used a simple learned CNN model with vector quantization for producing a final sequence of $1024$ discrete tokens.
    For details, see Appendix~\ref{app:models_datasets}.
    
    The base model $\model{1}$ (that we typically used as a source foundation model) had $24$ layers, $8$ heads and used the embedding dimension $d=512$.
    Our smaller models $\model{s}$ and $\proj{s}$ with the multiplier $s\in \msizes$ used the same architecture as $\model{1}$, but had a smaller embedding size $sd$.
    The number of trainable parameters in models $\model{1}$ through $\model{1/8}$ was approximately equal to $76.7$M, $19.5$M, $5$M and $1.3$M correspondingly.

    \paragraph{Weight Generators.}
    In our experiments, we used a specific choice of the MLP producing $\mat{P}$ and $\mat{Q}$ projection matrices.
    The input task identifier $\task$ was first linearly mapped to an $r$-dimensional vector (with $r$ typically chosen between $4$ and $32$), which after the \textsc{Swish} nonlinearity was followed by a linear layer producing diagonal elements of $\mat{P}$ and $\mat{Q}$ (as described in Sec.~\ref{sec:generators}).

    \subsection{Experimental Results}
        \def\modtwo{{\rm mod}\, 2}
\def\twohalves{1/2-1/2}

\subsubsection{Baseline Models}

    We started by pretraining a \gpt{} based autoregressive model $\model{1}$ on the image distribution ($\synth{}$ and $\imagenet{}$).
    We also trained a number of smaller baseline \gpt{} models $\model{s}$ for the target model multiplier $s\in \msizes$.
    
    Along with $\model{s}$ we also trained conditional autoregressive models $\cmodel{s}$.
    These models were identical to $\model{s}$, but were trained on sequences that in addition to image pixels also encoded the information about the requested task identifier $\task$ as the first embedding in the sequence.
    As a result, while $\model{s}$ is a model capable of generating arbitrary images, specialized models $\cmodel{s}$ can generate images with any predefined task identifier $\task$.
    The losses of all unconditional and conditional pretrained models are shown in Table~\ref{tab:synth-basic} for \synth{} and in Table~\ref{tab:imagenet-basic} for \imagenet{} datasets.
    In the following, we compare our specialized generated models $\proj{s}$ to corresponding {\em universal conditional} models $\cmodel{s}$.

\subsubsection{Projections from $\model{1}$}

    \begin{table}
  \small
  \begin{center} \begin{tabular}{c|cccc}
    $s$ & $\model{s}$ & $\cmodel{s}$ & $\proj{s}$ & $\jointproj$ \\
    \hline \\[-.6em]
    
    $1$ & $0.45$ & $\mathbf{0.39}$ & $(0.44)$ & $(0.46)$ \\
    $1/2$ & $0.51$ & $0.46$ & $\mathbf{0.39}$ & $\mathbf{0.39}$ \\
    $1/4$ & $0.62$ & $0.58$ & $0.47$ & $\mathbf{0.45}$ \\
    $1/8$ & $0.82$ & $0.78$ & $\mathbf{0.57}$ & $\mathbf{0.57}$ \\[.4em]
  \end{tabular} \end{center}
  \caption{
    Model losses on the \synth{} image dataset: (a) $\model{s}$ is a conventional autoregressive model using multiplier $s$; (b) $\cmodel{s}$ is an autoregressive model conditioned on $\task$ passed in the first input token; (c) $\proj{s}$ is a projected model co-training the original $\model{1}$ and task-specific $\proj{s}(\task)$; (d) $\jointproj$ is a single model simultaneously co-training $\model{1}$ and projections $\proj{s}(\task)$ for all $s\in \msizes$.
    Cells $s=1$ and $\proj{s}/\jointproj$ show ``source model'' $\model{1}$ performance after co-training (average value for all $s$ for $\proj{s}$).
  }
  \label{tab:synth-basic}
\end{table}

    As our first experiment, we used a pretrained $\model{1}$ to train our projected models $\proj{s}$.
    Specifically, we continued to tune $\model{1}$ and our projection operators (described in Sec.~\ref{sec:method}) using optimization objective~\eqref{eq:loss}.
    We carried out this process for each projected model size $s\in \msizes$ with the inner dimension $r=8$ (see Sec.~\ref{sec:models}).
    The number of parameters used by the projection operators ranged from $2.4$M for the $s=1/2$ model to $1.9$M for $s=1/8$ model.
    
    The results of our experiments are presented in Tables~\ref{tab:synth-basic} and~\ref{tab:imagenet-basic}.
    We observed that for both \synth{} and \imagenet{} generated projected models $\proj{s}({\task})$ typically outperformed universal conditional models since the former were generated from the task identifier $\task$ to only model task-specific knowledge $p(\vec{x}_1,\dots,\vec{x}_n|\task)$, whereas the latter had to approximate the entire $p(\vec{x}_1,\dots,\vec{x}_n,\task)$.
    The difference was especially pronounced on the \synth{} dataset, where performance of projected models was seen to be roughly comparable to the performance of universal conditional models that used $4$ times as many parameters\footnote{$\proj{1/8}$ roughly matching the performance of $\cmodel{1/4}$, $\proj{1/4}$ roughly matching $\cmodel{1/2}$, etc.}.
    On \imagenet{}, the gap was smaller and we estimated that the performance of $\proj{1/4}$ would roughly match the performance of a $1.6\times$ larger universal model, whereas $\proj{1/8}$ would roughly match a $2.5\times$ larger conditional \gpt{}.

    These results illustrate the promise of generating small specialized Transformers that proved to be much more suitable for modeling narrow image distributions (for a fixed $\task$, all \synth{} images share a similar background texture and \imagenet{} images are more likely to contain the same object or a scene).
    It is also worth noticing that the source model $\model{1}$ co-trained with projections ended up having the same perplexity as a standalone model suggesting that the resulting tuned $\model{1}$ can be used as a ``foundation model'' without any sacrifices to its performance.

\subsubsection{Co-Training Multiple Projected Models}

    In the followup experiment, we trained a model $\jointproj$ that simultaneously projected $\model{1}$ to all sizes $s\in \msizes$.
    Here, each $\proj{s}$ has it's own projection operator, while the source model $\model{1}$ needs to be compatible with all of the projected model sizes.
    Results presented in Table~\ref{tab:synth-basic} suggest that co-training projection models of various sizes did not lead to performance degradation (even showing minor improvement in some cases).
    As a result, the final tuned $\model{1}$ with projection operators for all $s\in \msizes$ provides us with the full spectrum of models: (a) a large foundation model describing the full image distribution, and (b) a capability to produce small task-specific models with a variety of different sizes.

\subsubsection{Additional Experiments}

    Appendix~\ref{app:extra_results} discusses additional experimental results with both \synth{} and \imagenet{} datasets.
    Specifically, we discuss our studies of (a) the effect of the source model size on the specialized model performance (\ref{app:extra-projections}); (b) cross-task knowledge transfer in projected models (\ref{app:extra-cross-task}) and (c) zero-shot generalization of projected models (\ref{app:extra-zero}).

\begin{table}
  \small
  \begin{center} \begin{tabular}{c|cc}
    $s$ & $\cmodel{s}$ & $\proj{s}$ \\
    \hline \\[-.6em]
    $1$ & $2.176$ & -- \\
    $1/2$ & ${\bf 2.211}$ & ${\bf 2.210}$ \\
    $1/4$ & $2.264$ & ${\bf 2.246}$ \\
    $1/8$ & $2.332$ & ${\bf 2.297}$ \\[.4em]
  \end{tabular}
  \end{center}
  \caption{Results for \imagenet{} with \simclr{} embeddings as task identifiers (evaluated on the validation set). The statistical error of $\cmodel{s}$ is around $0.001$ and of $\proj{s}$ is approximately $0.002$.}
  \label{tab:imagenet-basic}
\end{table}

\section{Discussion}
\label{sec:discussion}

    In this work, we propose and study a mechanism for converting the weights of a large Transformer into the weights of a smaller task-specialized model.
While in the ideal case scenario, the source Transformer is an arbitrary pretrained model, our technique involves tuning this source model.
We then show that the resulting ``projected'' task-specialized models outperform universal task-agnostic models and the improvement is more pronounced for smaller generated models and larger source models.
Interestingly, we can use a single source Foundational Model and a set of ``projection operators'' to generate a variety of specialized models of different sizes without any noticeable degradation of the source model performance.
We also demonstrate a cross-task knowledge transfer in projected models, when a wealth of training data for some tasks improves performance of projected models on related, but different tasks.

{\small
\bibliographystyle{icml2024}
\bibliography{refs}
}

\newpage
\appendix
\onecolumn

\section{Related Work}
\label{sec:related}

    {\bf HyperNetworks.}
One-shot generation of entire models has been popularized in \cite{HaDL16}.
Since then, multiple applications including those in image generation and segmentation \cite{Zhang2023Adding,Alaluf2022,Dinh2022,Nirkin2021,Kang2023scaling} (including conditioning Stable Diffusion \cite{Rombach2022High}), continual learning \cite{VonOswald2019continual}, 3D space representations \cite{Littwin2019,Sitzmann2020,Spurek2022hyperpocket} and network architecture search \cite{Zhang2018graph,Knyazev2021parameter} have been proposed.
In natural language processing with Transformers, HyperNetwork-based approaches have also been widely adopted \cite{Ye2021learning,Deb2022boosting,Tay2020hypergrid,Mahabadi2021parameter,Ivison2022hyperdecoders}.
Two most recent directions target task-dependent prompt generation \cite{He2022hyperprompt} and one-shot generation of model perturbations \cite{Phang23} given a single or few examples of a novel task.
Several other related approaches have been explored since then in \cite{Volk2023example,Ivison2022hint,Liang2023hart,Phang2024investigating,Zhao2024hypermoe,Li2024mend,Tack2024online,Mu2024learning}.

{\bf Generating smaller Transformer models.}
With the explosion of interest in Transformer models, there has been a growing interest in techniques for generating smaller Transformers from larger capable pretrained models.
For example, \cite{Xia2023sheared,Ma23,Chen2024streamlining} propose LLM pruning techniques and  \cite{Lin2020weight,Wang2023learngene,Xu2023initializing,Samragh23} outline approaches for initializing weights of a smaller model using a large Transformer model.
Two recent reviews \cite{Wang2024model,Tang2024survey} discuss recent publications exploring Transformer model pruning among other approaches.

\section{Weight Space Manifold}
\label{app:perspective}

    \def\defweights{\theta_\circ}
\def\task{t}
\def\prompt{r}
\def\prob{p}
\def\learnedw{\theta_*}
\def\seq{x}
\def\mW{\mat{W}}
\def\mL{\mat{L}}
\def\mR{\mat{R}}
\def\mS{\mat{S}}
\def\mapdown{\op{P}}
\def\mapup{\op{Q}}
\def\matdown{\mat{P}}
\def\matup{\mat{Q}}

Consider a Transformer-based LLM trained on rich input data distribution $\prob(\seq)$.
Given a specific task $\task$, one approach to solving it relies on choosing a prompt $\prompt(\task)$ and modeling task-specific distribution $\prob(\seq|\task)$ as $\prob_{\defweights}(\seq;\prompt(\task))$ with the prompt $\prompt$ used as a fixed sequence prefix.
Here the parameters $\defweights$ of the trained model are kept fixed and are independent of the task even if it is known in advance and does not change often.

Another way of specializing to task $\task$ is based on modifying the model itself.
Perhaps the most widely used approach is to fine-tune the model $\defweights\to \learnedw(\task)$ on a small number of demonstrations from the task-specific distribution $\prob(\seq|\task)$.
Note that the final weights $\learnedw(\task)$ generally depend on the tuning procedure, employed seeds, etc.

Yet another specialization technique is based on the HyperNetwork-based approach \cite{HaDL16}.
Here we assume that there exists an almost everywhere smooth {\em weight space manifold} $\theta(\task)$ such that for each task $\task\in T$ it approximately holds that $\prob_{\theta(\task)}(\seq) \approx p(\seq|\task)$.
This manifold could be chosen to approximate a set of pretrained fine-tuned models $\{\learnedw(\task_i)\}_i$ for some $\{t_i\in T\}_i$, or could be learned directly on a given distribution of tasks over $T$.
Note that by learning the manifold $\theta(\task)$, we store some information about the task-related knowledge in the model describing $\theta(\task)$ thus potentially improving the performance of the specialized generated models since they no longer need to carry information and knowledge relevant to other potential questions and queries.

\paragraph{Weight manifold.}
    The weight manifold of the model can be extremely complex.
    However, performing Taylor decomposition with respect to $\task$ around some fixed ``base task'' chosen here as $\task=0$, we can derive an approximate expression for $\theta(\task)$:
    \begin{gather}
        \label{eq:explicit}
        \theta(\task) \approx \theta(0) - \left( \frac{\partial^2 \mc{L}}{\partial \theta^2} \right)^{-1} \frac{\partial^2 \mc{L}}{\partial \theta \, \partial \task} \task,
    \end{gather}
    where $\mc{L}(\theta,\task)$ is a model loss and all derivatives are computed at $\task=0$ and $\theta=\theta(0)$.
    In other words, this equation explicitly defines the hyperplane tangent to the weight manifold at $\task=0$.
    However, performing this computation in practice can be exceptionally expensive and can only describe the local structure of the model.

\paragraph{Topology of the task space.}
    The structure of the weight manifold $\theta(\task)$ is defined by the model and the topology of the task space.
    In a typical setup, the task $\task$ could be, for example, an embedding of a conversation topic, some representation of the user writing style, image embedding, or metadata.
    If we choose $\task$ to be some normalized embedding $\xi(\seq)$, the task manifold becomes a sphere and $\theta(\task)$ could also have a similar topology.
    Given a normalized embedding $\xi$, we can also construct a distribution $\prob(\seq|\task)$ defined for all $\|\task\|\le 1$ as a linear interpolation between the marginalized distribution and $\prob(\seq|\xi)$ with $\|\xi\|=1$:
    \begin{gather*}
        \prob(\seq|\task) := (1 - \|\task\|) \int_{\|\xi\|=1} \prob(\seq|\xi) \prob(\xi) d\,\xi + \|\task\| \prob(\seq|\xi=\task/\|\task\|).
    \end{gather*}
    Using this constructed conditional distribution, the expansion around $\task=0$ becomes an expansion around the average distribution with the direction of $\task$ defining the embedding of samples that become more prevalent in the distribution as $\|\task\|$ grows.

\subsection{Learning the Weight Manifold}

    Performing the computation in Eq.~\eqref{eq:explicit} can be prohibitively expensive.
    An alternative approach to characterizing the weight manifold $\theta(\task)$ is then to learn this dependence.
    
    \paragraph{Linear approximation.}
    For example, if we are only interested in the local linear structure of $\theta(\task)$, we can rely on the approximation:
    \begin{gather}
        \label{eq:linear}
        \delta\theta_\alpha(\task) := \theta_\alpha(\task) - \theta_\alpha(0) \approx \sum_i \mS_{\alpha,i} t_i,
    \end{gather}
    where $\alpha$ is an index of a specific parameter.
    We can then learn both $\theta(0)$ and $\mS_{\alpha,i}$ by drawing samples from the distribution $\prob(\seq|\task)p(\task)$ and training the model to optimize the loss $\mc{L}(\theta(0),\mS) := \E_{(\seq,\task)\sim \mc{D}} L(\seq;\theta(\task))$.
    This approach requires $\dim \theta \cdot \dim \task$ total parameters to parameterize $\delta \theta(\task)$.

    \paragraph{Modular approximation.}
    The linear approximation can be trivially reduced to a form of a {\em modular network} used previously in multiple contexts.
    Here we can represent $\delta \theta$ via a linear combination of a large collection of individual ``modules'' $\mS_{\alpha,k}$:
    \begin{gather}
        \label{eq:modules}
        \delta \theta_\alpha = \sum_k \mS_{\alpha,k} \eta_k(\task),
    \end{gather}
    where $\eta_k(\task)$ can now be arbitrarily complex non-linear functions of $\task$.
    This modular approximation requires $\dim \theta \cdot \dim \eta$ parameters and could be ``lighter'' than the linear approximation if $\dim \eta < \dim \task$.
    Notice that if $\eta_k$ are linear functions, this is equivalent to a low-rank version of Eq.~\eqref{eq:linear}.

    \paragraph{Low-rank approximation.}
    The required number of parameters necessary to describe $\delta \theta$ can be reduced further by replacing kernels $\mW$ of all linear operators in the model with low-rank approximations:
    \begin{gather}
        \label{eq:lr}
        \delta \mW_\ell(\task) = \mW_\ell(\task) - \mW_\ell(0) \approx \mL_\ell(\task) \mR_\ell(\task)
    \end{gather}    
    with $\ell$ being the operator index and $\mL_\ell(\task) \mR_\ell(\task)$ forming a low-rank matrix.
    Both $\mL_\ell(\task)$ and $\mR_\ell(\task)$ can be represented as shallow MLPs that accept $\task$ as input and produce all matrix components.
    The required number of parameters to describe an $n\times m$ matrix $\delta \mW_\ell(\task)$ is now reduced to roughly $O((n+m)r\dim \task)$ with $r$ being the rank.
    In the simplest cast with $r=1$, $\mL_\ell$ is a column and $\mR_\ell$ is a single row.

\subsection{Projection Approximation}
\label{sec:proj}
    
    Another way of generating $\delta \theta_\alpha$ is motivated by noticing that the individual module matrices $\mS_{\ell,k}$ combined as in Eq.~\eqref{eq:modules} could sometimes be arranged to form a larger matrix.
    This can be viewed as a way of mapping a larger Transformer into a smaller model.
    
    One natural choice in this potentially very rich family of transformation comes from analyzing linear maps of the embedding space.
    Specifically, given two linear transformations $\mapup:\R^{ds} \to \R^{d}$ and $\mapdown:\R^{d} \to \R^{ds}$ mapping model activations between embedding spaces of two models with embedding sizes $d$ and $ds$ and $s=2^{-k} < 1$, $k\in \Z$, we can naturally define $\oper{s}$ via
    \begin{gather*}
        \oper{s}(\vec{z}) \equiv \mapdown (\oper{1}(\mapup \vec{z})),
    \end{gather*}
    where $\oper{s}$ and $\oper{1}$ are two linear operators in a smaller and a larger Transformer model correspondingly.
    This perspective gives rise to a family of weight generators at the core of our main approach discussed in Sec.~\ref{sec:method}.

    \paragraph{Connection to Modular Networks.}
    Conventional modular architectures can be viewed as a special case of the described linear weight transformation.
    Consider an $(dn)\times (dm)$ weight matrix $\mat{W}_1$ of $\oper{1}$ and a smaller $(dsn)\times (dsm)$ weight matrix $\mat{W}_s$ of $\oper{s}$.
    One can subdivide $\mat{W}_1$ into $(dsn)\times (dsm)$-sized blocks and choose
    \begin{gather*}
        V_{ijkl} = \sum_{\alpha=0}^{s^{-1}-1} \sum_{\beta=0}^{s^{-1}-1} r_{\alpha \beta} \delta_{k,i+dsn\alpha} \delta_{l,j+dsm\beta},
    \end{gather*}
    resulting in the weight $\mat{W}_s$ becoming a linear combination of these blocks.
    This particular form is frequently referred to as a {\em modular network} with individual blocks acting as {\em modules}.
    Here $\delta$ is a Kronecker delta and $r_{\alpha \beta}$ are $s^{-2}$ coefficients of individual modules forming $\mat{W}_s$.
    
\ifx\includeaux\undefined
\else
        For $\mapup(\vec{z})=\matup \vec{z}$ and $\mapdown(\vec{z})=\matdown \vec{z}$ with $\matup(t)$ and $\matdown(t)$ being two matrices, we simply get:
        \begin{gather*}
            \mat{W}_s = \matdown \mat{W}_1 \matup, \qquad \vec{b}_s = \matdown \vec{b}_1.
        \end{gather*}
        With a proper choice of $\matdown$ and $\matup$, we can, for instance, recover a low-rank approximation \eqref{eq:lr}, or a combination of modules similar to that in \eqref{eq:linear}.
    
    \subsection{Mapping a Large Transformer to a Small Transformer}
    
        The approach outlined in Section~\ref{sec:proj} allows us to parameterize the weight manifold of a smaller model using fixed matrices $\mat{W}_1$ of a larger Transformer model.
        And while there it can be interpreted as performing projections on the model embedding spaces, there is no immediate connection between the functions of these two models.
        In our analysis, we could not identify a natural way of mapping a large pretrained Transformer model into the weight manifold of a smaller model.
        
        However, there exists a way of defining this correspondence by restricting the architecture of a large model and modifying the training scheme.
        To see this, we represent the activations of a large Transformer model as $(\vec{z},\hat{\vec{z}})$ with $\vec{z} \in \R^{ds}$ and $\hat{\vec{z}} \in \R^{d(1-s)}$.
        A kernel $\mat{W}$ of any linear layer of a large Transformer can then be represented in a block form $\mat{W}=((\mat{W}_{00},\mat{W}_{01}),(\mat{W}_{10},\mat{W}_{11}))$ resulting in the following linear transformation:
        \begin{gather*}
            \vec{z}_{\rm out} = \mat{W}_{00} \vec{z} + \mat{W}_{01} \hat{\vec{z}}, \quad \hat{\vec{z}}_{\rm out} = \mat{W}_{10} \vec{z} + \mat{W}_{11} \hat{\vec{z}}.
        \end{gather*}
        
        By choosing $\mat{W}_{10}=\mat{W}_{11}=0$ in the key and query matrices, we can guarantee that the attention will be defined only by the $\vec{z}_{\rm out}$ components.
        The transformation of $\vec{z}$ itself can then be viewed as a \dots
    
    \subsection{Representation Learning}
    
        Our approach can also be viewed as a representation learning method.
        The idea is to train a feature extractor $\xi(x)$ and generate $t=\xi(x) \cdot s$, where $s$ is a random scalar variable on $[0,1]$.
        We then minimize the autoregressive loss for a sample $x$ given this randomized task ID $t$.
        Minimization of the average loss achieves two things: fits our family of models to the emergent $p(x|t)$ and also finds a parameterization $\xi(x)$ reducing the error of this model.
\fi

\section{Models and Datasets}
\label{app:models_datasets}

    \subsection{Training Details}

A typical model has been trained for $200{\rm k}$ to $800{\rm k}$ steps with Adam optimizer and a learning rate of order of $10^{-3}$ with cosine learning rate decay schedule ($10{\rm k}$ warmup steps).
We typically used very weak weight decay ($10^{-10}$ to $10^{-8}$), but our dropout was frequently set to $10\%$.

\subsection{VQ-VAE Model}

We trained a separate VQ-VAE model \cite{VanDenOord2017neural} for mapping $64\times 64$ RGB images to sequences of $1024$ tokens taking values in a discrete set of size $512$.
Model encoder contained $3$ CNN layers with $3\times 3$ kernels and leaky ReLU nonlinearities.
Layers had the following depths and strides: $(32, 2)$, $(64, 1)$, $(64, 1)$.
The decoder was composed of $4$ transpose-convolutional leaky ReLU layers.
Layers had the following depths, kernel sizes and strides: $(64, 3, 2)$, $(64, 3, 1)$, $(32, 3, 1)$ and the final $(3, 1, 1)$.
The full VQ-VAE model was typically trained with $L_2$ loss and $\beta=0.2$ \cite{VanDenOord2017neural}.

\subsection{\synth{} Dataset}

Each \synth{} image was generated by (a) first using $\task$ to produce an image texture and (b) overlaying one of the MNIST images on top of this texture.
Examples of generated images can be found in Fig.~\ref{fig:sample}.

\paragraph{Texture.}
Task identifier $\task$ was sampled uniformly from $[0,1]^{18}$ and contained information about the image affine transformation like angle of rotation ($t_0$), scale ($t_1$), rotation center $x$ and $y$ coordinates ($t_2$ and $t_3$).
The actual angle was equal to $2\pi t_0$, the scale was chosen as ${\rm sc}(t_1) := 2 + 18 t_1$.
The image was also ``warped'' by applying a transformation $x\to x + a_x \cos(s_x x)$ and $y\to y + a_y \cos(s_y y)$ with $a_{x,y} := {\rm relu}(\bar{a}_{x,y} - \alpha) / (1 - \alpha)$, $\alpha=0.3$, $\bar{a}_x=t_4$, $\bar{a}_y=t_5$ and $s_x=s_y={\rm sc}(t_6)$.
Each color component (red, green, blue) of the image was shifted relative to others as defined by $t_7$ for R, $t_8$ for G and $t_9$ for B.
The image texture itself was randomly chosen from one of $3$ classes (depending on which of $t_{10}$, $t_{11}$ and $t_{12}$ is larger) based on the following function profiles: (a) $\cos \tilde{x}$ with $\tilde{x}$ being a transformed $x$ coordinate, (b) $\cos \tilde{\rho}$ with $\tilde{\rho}$ being a distance to the origin in transformed coordinates, (c) $\cos (10 g t_{13} + 2 t_{14})$ with $g:=\cos(\tilde{x}/3)\cos(\tilde{y}/3)$ and $\tilde{y}$ being a transformed $y$ coordinate.
Finally, the generated texture was multiplied by a random 3-channel RGB-color vector $(\beta+\gamma t_{15},\beta+\gamma t_{16},\beta+\gamma t_{17})$, where we chose $\beta=0.5$ and $\gamma=1.5$.
The produced image was clipped to a $[-1,1]$ output range.

\paragraph{Digit.}
In the training dataset split, we used a random augmentation by horizontally flipping digit images with $50\%$ probability.

\section{Additional Experimental Results}
\label{app:extra_results}

    \subsection{Projections from $\model{s}$}
    \label{app:extra-projections}

    The task-specialized projected model $\proj{s}(\task)$ is expected to derive its capabilities from: (a) the source model $\model{1}$, the weights of which are used to generate $\proj{s}$ and (b) the projection operator parameterizing the way that these source weights are used depending on the task identifier $\task$.
    It is interesting to ask which of these sources is more important for the performance of the projected model $\proj{s}$.
    
    We explored this question by using source models $\model{s}$ of different sizes $s$ for producing projected models $\proj{s'}$ with $s'<s$.
    We also varied the size of the inner dimension $r$ to control the number of parameters in the projection operator itself.
    The number of parameters for the source $\model{s}$ with a particular value of $r$ is roughly proportional to $rs$.
    Projecting from $\model{1}$ with $r=8$ is thus roughly equivalent to projecting from $\model{1/2}$ with $r=16$ and from $\model{1/4}$ with $r=32$.
    The results of our experiments presented in Table~\ref{tab:varying-size} suggest that the performance of the projected model is generally better for larger source models.
    However, the benefit from using larger source models may be less pronounced once they exceed a particular size.

    \begin{table}[b]
  \small
  \begin{center} \begin{tabular}{c|c|cc|ccc}
    & $\model{1}$ & \multicolumn{2}{c|}{$\model{1/2}$} & \multicolumn{3}{c}{$\model{1/4}$} \\
    \hline
    $r=$ & $8$ & $8$ & $16$ & $8$ & $16$ & $32$ \\
    \hline
    $\model{1/4}$ &	$\mathbf{0.47}$ & $0.50$ & $0.49$ & -- & -- & -- \\
    $\model{1/8}$ & $\mathbf{0.57}$ & $0.59$ & $\mathbf{0.57}$ & $0.71$ & $0.70$ & $0.67$
  \end{tabular} \end{center}
  \caption{
    Training losses for $\model{1/4}$ and $\model{1/8}$ models (rows) projected from $\model{1}$, $\model{1/2}$ and $\model{1/4}$ source models (columns) with different values of the projector inner dimension $r$.
    The projection operator size for the source $\model{1}$ with $r=8$ ($\sim 2M$ parameters) approximately matches the projection operator sizes for $\model{1/2}$ with $r=16$ and $\model{1/4}$ with $r=32$.
  }
  \label{tab:varying-size}
\end{table}

\subsection{Cross-Task Knowledge Transfer}
    \label{app:extra-cross-task}

    \begin{table}
  \small
  \begin{center}
  \begin{tabular}{cccc}
    & $1$ & $1/8$ & $1/8$ and $1$ \\
    \hline \\[-.6em]
    $\twohalves$ & ${\bf 2.382}$, ${\bf 2.040}$ & $2.395$, $2.052$ & $2.392$, $2.045$ \\
    $\modtwo$ & ${\bf 2.221}$, ${\bf 2.201}$ & $2.233$, $2.214$ & $2.225$, $2.206$ \\[.4em]
  \end{tabular}
  \end{center}
  \caption{
    Losses on $C_1$ and $C_2$ of task-specific projected \gpt{} models with $s=1/2$ computed on the \imagenet{} validation set ($50k$ samples; the estimated error is approximately $0.002$).
    The rows correspond to two different task partitions across $C_1$ and $C_2$.
    Three columns show results for different sizes of the training sets for tasks in $C_1$ and $C_2$: (a) all tasks use all available training samples; (b) all tasks use $1/8$ of available training samples; (c) tasks in $C_1$ are trained with $1/8$ of available training samples and tasks in $C_2$ use all training samples.
  }
  \label{tab:transfer}
\end{table}

    In our approach, models generated for different tasks indirectly share their weights via the source model.
    It is thus interesting to explore the degree to which these models exchange information and quantify knowledge transfer across different tasks.
    We accomplish this by noticing that providing additional training samples for some tasks may boost the performance of models generated for related but different tasks.
    We study this phenomenon in \imagenet{} models by subdividing the training set into two parts based on the image labels: (a) part $C_1$ contains images with labels below $500$ and $C_2$ with labels above $500$ and (b) part $C_1$ containing images with labels $l \equiv 0 \, (\modtwo)$ and part $C_2$ with image labels $l \equiv 1 \, (\modtwo)$.
    We denote the first scenario as ``$\twohalves$'' and the second scenario as ``$\modtwo$''.
    Since \imagenet{} labels are not entirely random, but are ordered in a semantically meaningful way, we expect that tasks in $C_1$ and $C_2$ sets in the ``$\modtwo$'' scenario are more closely related to each other compared to the ``$\twohalves$'' setup.

    In our experiments, we trained three $\proj{1/2}$ models for each of the two scenarios: (a) using all training samples for both $C_1$ and $C_2$; (b) using only $1/8$ of all existing training samples for both $C_1$ and $C_2$; (c) using $1/8$ of training samples for $C_1$ and all training samples for $C_2$.
    The results of our experiments are presented in Table~\ref{tab:transfer}.
    We observed that providing additional examples for tasks in $C_2$ actually boosts the performance of models generated for tasks in $C_1$.
    This boost is especially noticeable in the ``$\modtwo$'' scenario, where tasks in $C_1$ and $C_2$ are closely related.
    Similarly, notice that the performance on $C_2$ suffers from a smaller training set for $C_1$.
    In other words, there is a noticeable interaction and knowledge transfer between models generated for different, but related tasks. 

\subsection{Zero-Shot Generalization}
    \label{app:extra-zero}

\begin{figure*}[h]
      \centering
      \includegraphics[width=0.96\textwidth]{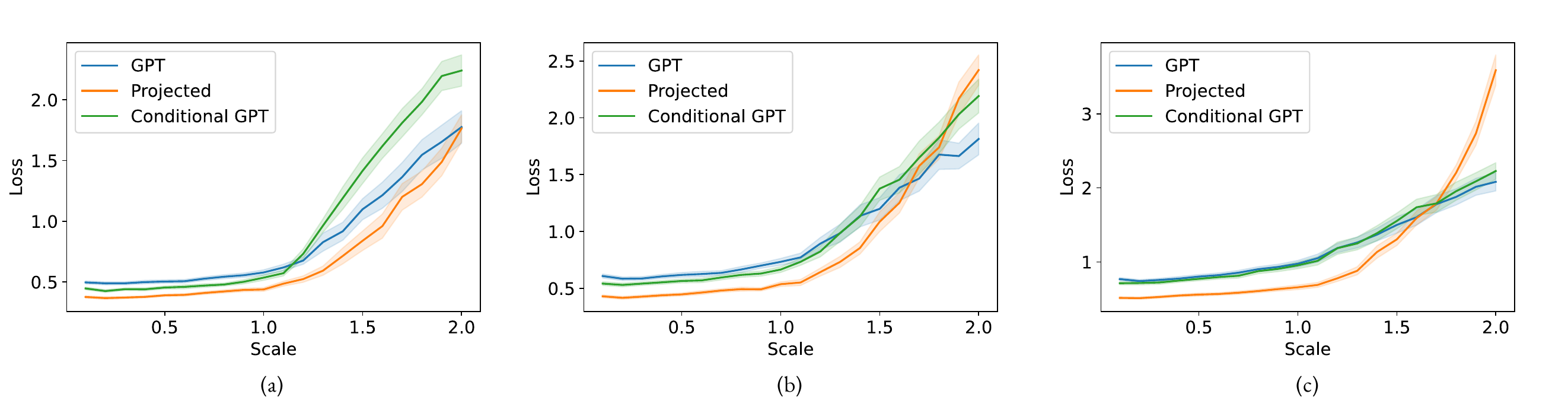}
      \caption{
        Average loss as a function of the scale component $\task_{\rm scale}$ (plotted on the $x$ axis) of the task identifier $\task$ with the following projected models: (a) $\proj{1/2}$, (b) $\proj{1/4}$, (c) $\proj{1/8}$.
      }
      \label{fig:scale}
    \end{figure*}

    An important property of any model specialization technique is it's zero-shot generalization capability.
    We explored model generalization for \synth{} dataset by varying the scale component $\task_{\rm scale}$ of the task identifier $\task$ from $0$ to $2$, while the training range of scales was $[0,1]$.
    In Figure~\ref{fig:scale} we show the average loss of different models as a function of $\task_{\rm scale}$ (horizontal {\em scale} axis in all of the plots).
    Notice that while all models start degrading around $\task_{\rm scale}=1$, the projected model $\proj{1/2}$ shows better generalization than the Conditional \gpt{} model $\cmodel{1/2}$.
    However, $\proj{1/4}$ and especially $\proj{1/8}$ have much greater difficulty generalizing to $\task_{\rm scale} \approx 2$, which could be explained by the fact that the corresponding projection operators used a large number of parameters compared to the projected model size.

\end{document}